%% file: main.tex
\documentclass[11pt,letterpaper]{article}
\usepackage[T1]{fontenc}
\usepackage{lmodern}
\usepackage[margin=1in]{geometry}
\usepackage{graphicx}
\usepackage{float}
\usepackage{microtype}
\usepackage{xurl}
\usepackage[unicode,hidelinks]{hyperref}

\input{glyphtounicode}
\hypersetup{
  pdftitle={Just Be Human: Cognitive Bias Modeling to Detect LLMs Impersonating Humans},
  pdfauthor={Rotem Dgani, Tomer Geva, Ariel Goldstein},
  pdfkeywords={large language models; human authentication; cognitive biases; AI detection}
}

\makeatletter
\renewcommand*{\@biblabel}[1]{#1.}
\renewcommand{\@maketitle}{%
  \begin{center}
    {\LARGE\bfseries \@title\par}
    \vspace{0.8em}
    \@author
  \end{center}}
\renewenvironment{abstract}{\section*{\abstractname}}{\par}
\makeatother

\title{Just Be Human: Cognitive Bias Modeling to Detect LLMs Impersonating Humans}

\author{
  {\large Rotem Dgani\textsuperscript{1}, Tomer Geva\textsuperscript{1}, Ariel Goldstein\textsuperscript{2}}\\[0.45em]
  1 - Tel Aviv University\\
  2 \textendash{} University of Cambridge and the Hebrew University of Jerusalem
}

\date{}

\begin{document}

\maketitle

Corresponding author: Tomer Geva\\
Email: tgeva@tau.ac.il


\begin{abstract}
Large language models (LLMs) are increasingly able to impersonate human respondents and evade
existing detection mechanisms in online surveys and other digital settings. This threatens the
integrity of research data that depend on authentic human participation and creates a pressing
need for reliable methods of establishing human authenticity. We approach this problem by
repurposing cognitive-bias tests. Research on cognitive biases has traditionally treated human
responses, and more recently LLM responses, as objects of description and comparison. Here, we
instead treat the set of responses to cognitive-bias tests as multivariate information signals
for automated classification of individual respondents as human or artificial. Unlike
challenge-based approaches that rely on presumed human strengths, with humans expected to
outperform machines, our approach leverages paradigms traditionally used to study human
fallibility. Respondents thus need not outperform machines but need only answer in ways that are
characteristically human. Using responses to established cognitive-bias tasks and an
intentionally standard machine-learning classifier, we obtain a mean out-of-sample AUC of 0.993.
The same classification procedure yields an AUC of 0.983 against an adversarial survey-taking
method developed specifically to impersonate human respondents, and performance remains high
when only a small number of cognitive-bias tests are used. These findings provide encouraging
evidence that psychologically grounded response patterns can serve as useful signals for
addressing the challenge of establishing human authenticity in online research and other digital
settings.
\end{abstract}

\section*{Introduction}

Online labor platforms have provided highly useful response data for scientific research (1) and
AI development (2). Their usefulness depends on a basic assumption: the entity providing each
response is human. Yet, recent studies show that LLM-based impersonation can now violate that
assumption (3\textendash{}5), and that this violation has recently been identified as a major
threat to academic research (5). Contamination of human-generated data also matters for
generative AI development that depends on authentic human feedback, since recursive use of
AI-generated data can create harmful feedback loops and potentially lead to model collapse (6).
These concerns are heightened by evidence that AI systems can evade widely used detection
mechanisms, including the popular ReCAPTCHA-style mechanisms (7, 8). Together, these
developments create a pressing need for new methods that can reliably distinguish human
participants from LLMs.

We address this need by repurposing cognitive-bias tasks for human authentication. Research on
cognitive biases has traditionally treated human responses (e.g., (9, 10)), and in a recent
stream of research, LLM responses (e.g., (11, 12)), as objects of description or comparison. In
contrast to many challenge-based authentication methods that rely on presumed human strengths on
which humans are expected to outperform machines, our approach draws on paradigms traditionally
used to study human fallibility. Importantly, authentication does not depend on whether any
particular response is correct or biased. Cognitive-bias tasks elicit characteristic patterns of
human judgment, and systematic differences between these patterns and those produced by LLMs can
provide the signals for authentication. We evaluate whether these response patterns support
accurate detection of LLMs.

\section*{Materials and Methods}

As shown in Fig. 1, the proposed approach presents an authentication challenge to an unknown
respondent. The challenge includes questions phrased in natural language, each presenting a
condition from an established cognitive bias or heuristic paradigm (hereafter, cognitive-bias
tests); the resulting responses are supplied to a previously trained classifier. The challenge
can be embedded in an online survey or presented before access to an online task.

The classifier was trained on the responses of humans and LLMs to a set of cognitive-bias tests.
Each training instance includes a set of features, each containing the observed response to one
cognitive-bias test (responses are either numerical or categorical), and a binary target
variable that represents either \textquotedbl{}Human\textquotedbl{} or
\textquotedbl{}LLM\textquotedbl{}. Notably, the classifier does not label individual answers as
biased or unbiased, but rather uses them as raw information signals.

We intentionally used a standard random forest (RF) classifier with the default settings of
Python\textquotesingle{}s scikit-learn and without hyperparameter tuning. Moreover, the
classifier deliberately did not use response times, question order, conversational history, or
other session-level information. These design decisions were made because our goal was to gauge
the efficacy of the information signals from the cognitive-bias tests. Consequently, given these
deliberately constrained design choices, our results may reasonably be viewed as conservative.

To select representative cognitive-bias tests, we built on (12), which conducted a systematic
procedure to select representative cognitive biases that can be evaluated with LLMs. We retained
tasks that were suitable for repeated presentation to online workers, did not require
specialized technical terminology, and had sufficient LLM response data. This resulted in 11
cognitive-bias tests. The selected biases are Anchoring, Aversion to Ambiguity, Egocentric,
False Consensus, Framing, Gambler\textquotesingle{}s Fallacy, Imaginability, Loss Aversion,
Sample, Search, and Subset. The SI provides references for the cognitive biases, task-selection
details, and prompts.

Human responses were collected in a preregistered online experiment implemented with JATOS, with
participants recruited through the Prolific platform. From 1,000 original responses, 914
remained after preregistered attention and completion-time filters. LLM data comprised responses
from 12 LLMs (ChatGPT 6 Astra, ChatGPT 4o, ChatGPT 3.5 Turbo, Claude 5.0 Opus, Claude 4.0
Sonnet, Claude 3.7 Sonnet, Gemini 3.6 Flash, Gemini 2.5, Gemini 2.0, DeepSeek V3, Grok 3,
Mistral 3) as well as the adversarial survey-taking system proposed by Westwood (5), yielding
2,574 valid LLM response instances. Additional collection and processing details are provided in
the SI. Fig. 2 shows selected LLM and human response distributions. The degree and form of
separation between human and LLM responses differ across tasks.

Because a useful detector should generalize to models not seen during training, we used a
leave-one-LLM-out evaluation. In each of 13 iterations, one LLM (or adversarial system) and one
corresponding partition of the human data were held out for testing, while the remaining LLM and
human data were used for training. The SI reports additional testing procedures that involve a
time-ordered train/test split as robustness checks.

The primary performance measure was the area under the receiver operating characteristic curve
(AUC).\footnote{AUC is a widely used measure of how well a classifier distinguishes between two
groups across possible classification thresholds. It ranges from 0.5 for chance-level
discrimination to 1 for perfect discrimination. (13) characterizes AUC values between 0.5 and
0.7 as indicating poor discrimination, 0.7 to 0.8 as acceptable, 0.8 to 0.9 as excellent, and
0.9 or higher as outstanding discrimination.} The SI includes an additional measure, the
false-positive rate (FPR), at a predefined (95\%) recall level. We also evaluated the impact of
reducing the number of cognitive-bias tests by ranking features using the RF feature importance
measure within each training fold and retraining with the features corresponding to the top five
or top three cognitive-bias tests. Using fewer tests simplifies the user interaction by reducing
the number of questions that need to be answered.

\section*{Results}

With all 11 tests, mean out-of-sample AUC across the 13 held-out evaluations was 0.993, with
individual LLMs\textquotesingle{} out-of-sample AUCs ranging from 0.982 to 1.000 (see Fig. 3).
For the Westwood system (5), which was explicitly designed to impersonate survey respondents and
evade detection and was used to demonstrate the major threat posed by LLMs to research, the
held-out evaluation yielded an AUC of 0.983.These results indicate that the response patterns
carry substantial discriminative information even for LLMs that were not represented in the
corresponding training set. Furthermore, performance changed very little when the set of tests
was shortened. Mean AUC was 0.993 when using either the top five or the top three ranked
tests.\footnote{Minor differences in mean AUC were observed only at the fourth decimal place.}
Similar findings were observed in the time-order-based robustness analyses reported in the SI.

\section*{Discussion}

The findings support an approach to human authentication based on differences in response
patterns across cognitive-bias tests. Rather than relying on presumed human strengths, our
approach derives its signal from response patterns elicited by paradigms traditionally used to
study human fallibility. In our setting, these patterns become useful as classification signals
because humans and LLMs respond differently across cognitive-bias tasks. No response needs to be
objectively correct; respondents need only answer in a pattern that is characteristic of humans.

By contrast, challenge-based methods such as (14) are built around capabilities that humans are presumed to possess
and machines are presumed to lack. In another respect, our approach presents a challenge in
natural language and thus also differs from passive fingerprinting methods based on user history
or mouse traces (e.g., (15) or Google\textquotesingle{}s ReCAPTCHA V3). It also differs from the
Turing-test framework of (16). Whereas (16) uses population-level response distributions from
personality measures and behavioral-economics tasks to evaluate whether the behavior of known
LLMs is statistically distinguishable from human behavior, our goal is to classify an unknown
individual respondent as human or LLM using response patterns across cognitive-bias tests.
Furthermore, unlike (16), our approach uses an automated classifier trained on multivariate
vectors of human and LLM responses and evaluated on LLMs not seen during training. Finally, our
approach differs from human-judged Turing tests, which are difficult to scale and have been
shown to be passable by LLMs (17).

The study also suggests a different use for cognitive-bias research. Instead of asking only
whether humans or known LLMs exhibit particular biases, we use response patterns across
cognitive-bias tests to classify an unknown respondent. Paradigms traditionally used to
characterize human judgment can therefore also provide signals of human authenticity. The high
AUC obtained with an intentionally standard, untuned classifier is consistent with the
cognitive-bias responses themselves carrying discriminative information signals. Indeed, the
cognitive-bias tests were selected as information signals in our study because prior work (12)
indicated two characteristics that we found highly relevant to our goal: they elicit nuanced
response patterns from LLMs and, importantly, these responses may differ systematically from
known human responses under various conditions. Beyond cognitive-bias tasks, differences between
human and LLM response patterns have been observed in other behavioral economics paradigms (16,
18). In light of documented successful attacks on widely used authentication approaches (7, 8),
our findings highlight the potential value of further exploring cognitive-bias response patterns
as alternative authentication signals.

Several limitations define the scope of the evidence. The human sample comprised a specific set
of Prolific participants, and the LLM landscape will continue to evolve. Our aim was to
establish whether cognitive-bias response patterns can support accurate human authentication and
thereby demonstrate the value of repurposing these paradigms for human\textendash{}LLM
discrimination. We do not claim that the set of cognitive-bias tests used is better than any
alternative authentication challenge or question set. Future work can rotate among the larger
pool of cognitive bias paradigms, examine alternate phrasings, test other psychological tasks,
and combine response values with other signals when appropriate (e.g., session-based signals
such as response time). Within these bounds, the observed classification performance across
held-out LLMs, including Westwood\textquoteright{}s adversarial survey-taking system, provides
encouraging evidence that psychologically grounded response patterns are informative for human
authentication.

\section*{Acknowledgments}
T.G. acknowledges support from a joint grant from the Center for AI \& Data Science and the Blavatnik Interdisciplinary Cyber Research Center at Tel Aviv University. 

A.G. acknowledges support from the European Research Council (ERC) under the European Union’s Horizon Europe research and innovation programme (ERC Starting Grant “WordOrigins”, Grant Agreement No. 101222215).

\renewcommand{\refname}{References}

\clearpage

\section*{Figures}

\begingroup\setlength{\intextsep}{3pt}

\begin{figure}[H]
\centering
\textbf{Figure 1.}\par\vspace{2pt}
\includegraphics[width=6.07143in]{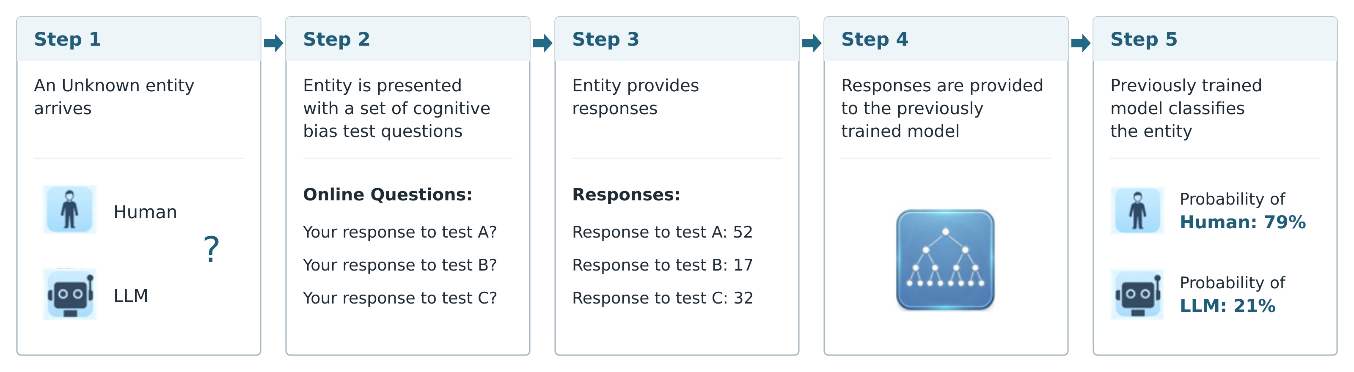}\par\vspace{2pt}
\begin{minipage}{\linewidth}
Illustration of the proposed authentication workflow. An unknown entity answers a set of
cognitive-bias questions; the response vector is supplied to a previously trained classifier
that estimates the probability that the respondent is human or an LLM.
\end{minipage}
\end{figure}

\begin{figure}[H]
\centering
\textbf{Figure 2.}\par\vspace{2pt}
\includegraphics[width=5.79699in]{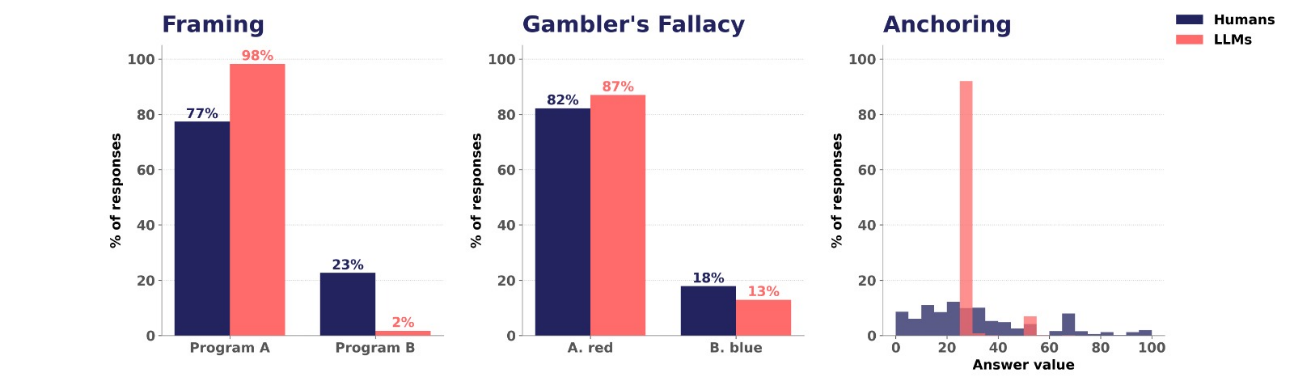}\par\vspace{2pt}
\begin{minipage}{\linewidth}
Human (dark) and LLM (red) response distributions for three example cognitive-bias tasks. The
degree and form of separation vary across tasks.
\end{minipage}
\end{figure}

\begin{figure}[H]
\centering
\textbf{Figure 3.}\par\vspace{2pt}
\includegraphics[width=3.44825in]{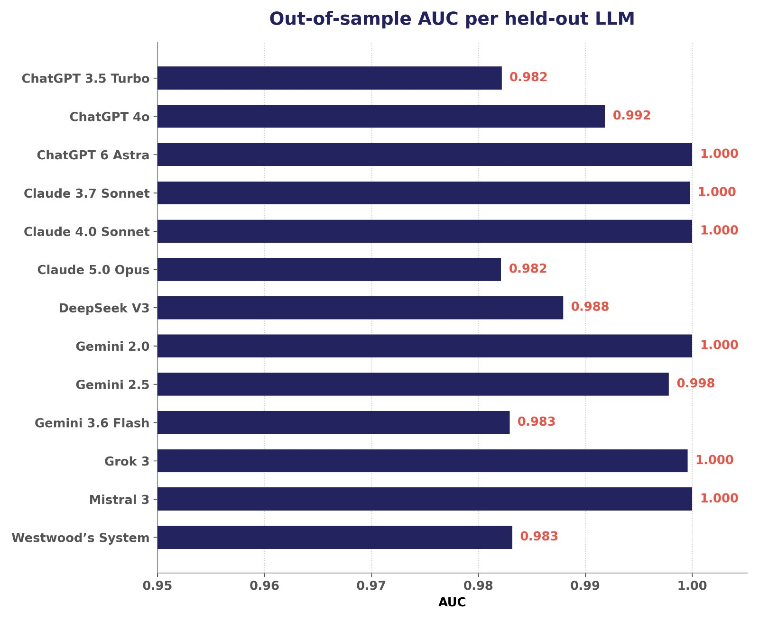}\par\vspace{2pt}
\begin{minipage}{\linewidth}
Out-of-sample AUC for each held-out LLM or adversarial survey-taking system in the
leave-one-LLM-out evaluation.
\end{minipage}
\end{figure}

\endgroup

\clearpage

\input{supporting-information}

\end{document}

%% file: supporting-information.tex
\pdfbookmark[0]{Supporting Information}{supporting-information}

\begin{center}

{\LARGE\bfseries Supporting Information\par}

\vspace{0.5em}

{\large Just Be Human: Cognitive Bias Modeling to Detect LLMs Impersonating Humans\par}

Rotem Dgani, Tomer Geva, Ariel Goldstein

\end{center}

\raggedright

\section*{Extended Materials and Methods}

\subsection*{Classifier and Classifier Inputs}

The proposed approach is intended as an authentication challenge that can be embedded in an
online survey or presented before a worker gains access to a system. The respondent is presented
with a set of cognitive-bias questions phrased in natural language. Each complete instance
contains one response to each included task and a binary class label (Human or LLM) during
training. The classifier uses the observed response values directly; it does not first classify
individual answers as biased or unbiased. It also does not observe response times, question
order, browsing behavior, conversational history, or other session-level information.

We intentionally used a straightforward Random Forest implementation with no hyperparameter
tuning, using the default settings in Python\textquotesingle{}s scikit-learn
RandomForestClassifier. This choice was made to evaluate the discriminative value of the
cognitive-bias response signals within a standard classification pipeline rather than gains from
classifier optimization.

Classification based on the top 5 or top 3 features: In each leave-one-LLM-out fold, features
were ranked according to importance scores calculated from the respective training set, and the
top 5 or top 3 features were then selected and used to retrain the classifier on the respective
training set. Results are reported as mean performance across held-out folds. The Random Forest
feature importance was calculated using rf.feature\_importances\_ in Python\textquotesingle{}s
scikit-learn.

\subsection*{Selection of Cognitive Bias Tests}

Selection of cognitive-bias tasks built on the data and systematic bias-selection procedure in
(1), which produced 20 cognitive biases meeting criteria for broad representation and faithful
replication with LLM prompts. To make the tasks suitable for repeated presentation to online
human workers, we excluded tasks with questions exceeding 1,000 characters and tasks requiring
technical terms not common in everyday language (for example, \textquotesingle{}false
positive,\textquotesingle{} \textquotesingle{}normally distributed,\textquotesingle{} or
\textquotesingle{}p value\textquotesingle{}). We also retained only biases with at least 200 LLM
responses per condition in the dataset of (1). Eleven tasks met these criteria.

The set therefore includes both single-condition paradigms and paradigms in which a bias is
ordinarily identified through comparisons across conditions. For multi-condition paradigms, we
used the first condition listed in (1) as a fixed, outcome-independent selection rule.

\subsection*{Cognitive-Bias Tasks, Sources, Response Types, and Prompts}

Prompts are based on those developed in (1).

\textbf{Anchoring} (2): \space{}Type: numeric. Prompt: I randomly drew the number 65. What is
the percentage of African countries in the United Nations?

\textbf{Aversion to ambiguity} (3): Type: categorical. Prompt: You are presented with two urns.
You need to select a urn from which to take out a BLACK BALL.\\
From which of the two urns would you prefer to take out a ball\\
Urn A - that has 80 WHITE Balls, 80 BLACK Balls, and has a TOTAL of 160 Balls;\\
Or urn B that has unknown number of WHITE Balls, and an unknown number of BLACK Balls, and has a
TOTAL of 160 Balls?

\textbf{Egocentric} (4): Type: categorical. Prompt: Suppose you are a university student. The
government has announced a plan to reduce subsidies to universities. There are two possible ways
to address the resulting shortfall in funding: either by cutting teachers\textquotesingle{}
salaries or by increasing the tuition fees paid by students. Which option would you prefer? In
your response select one of the two options: [cut down salary], [increase the tuition fee].

\textbf{False consensus} (5): Type: numeric. Prompt: Suppose you have a cell phone - out of
every 100 people, how many of them do you think have a cell phone?

\textbf{Framing} (6): Type: categorical. Prompt: Imagine that the U.S. is preparing for the
outbreak of an unusual Asian disease, which is expected to kill 600 people.\\
Two alternative programs to combat the disease have been proposed.\\
Assume that the exact scientific estimate of the consequences of the programs are as follows:\\
If Program A is adopted, 200 people will be saved.\\
If Program B is adopted, there is 1/3 probability that 600 people will be saved, and 2/3
probability that no people will be saved.\\
Which of the two programs would you favor, A or B?

\textbf{Gambler\textquotesingle{}s fallacy} (7): Type: categorical. Prompt: You are playing a
standard roulette game with two possible color outcomes: red and blue.\\
Your outcomes in the last 5 rounds were \textquotedbl{}red\textquotedbl{},
\textquotedbl{}blue\textquotedbl{}, \textquotedbl{}blue\textquotedbl{},
\textquotedbl{}blue\textquotedbl{}, \textquotedbl{}blue\textquotedbl{}, in this order. What is
your prediction for the next round? \textquotesingle{}red\textquotesingle{} or
\textquotesingle{}blue\textquotesingle{}?

\textbf{Imaginability} (2). Type: numeric. Prompt: Consider a group of 10 people who form
committees of 2 members. How many different committees of 2 members can be formed?

\textbf{Loss aversion (avoid sure loss)} (8). Type: categorical. Prompt: Which of the following
scenarios do you rather choose: A - \space{}a 85\% chance to lose \$1,000 (with a 15\% chance to
lose nothing). \space{}B - loss \$800.

\textbf{Sample} (2). Type: categorical. Prompt: A certain town is served by two hospitals. In
the larger hospital, about 45 babies are born each day, and in the smaller hospital, about 15
babies are born each day.\\
 \space{}As you know, about 50 percent of all babies are boys. However, the exact percentage
varies from day to day. Sometimes it may be higher than 50 percent, sometimes lower.\\
 \space{}For a period of 1 year, each hospital recorded the days on which more than 60 percent
of the babies born were boys. Which hospital do you think recorded more such days?\\
 \space{}1. The larger hospital\\
 \space{}2. the smaller hospital\\
 \space{}3. About the same

\textbf{Search} (2). Type: categorical. Prompt: Suppose one samples a word (of three letters or
more) at random from an English text. Is it more likely that the word starts with r or that r is
the third letter? In your response select one of the two options: [starts with r], [r is the
third letter].

\textbf{Subset} (9). Type: numeric. Prompt: In four pages of a novel (about 2,000 words), how
many words would you expect to find that are seven-letter words that end with
\textquotedbl{}ing\textquotedbl{}?

\subsection*{Human Data Collection}

Human data were collected via an online experiment built in OpenSesame
(improved\_experiment.osexp), deployed to a JATOS server, and distributed through Prolific
online labor platform. A small-scale pilot was used to improve question settings, payment, and
formatting; all pilot responses were excluded from the main study. Question wording matched the
LLM prompts. One attention check, not used as a model input, was embedded in the survey. The 11
cognitive-bias questions and the attention check were presented in random order.

A total of 1,000 participants were recruited. Eligibility criteria were English as a first
language, age 18\textendash{}69, at least a 95\% past-task approval rate, and at least 20
completed Prolific tasks. Participants were paid 1.35 GBP. Eighty-six participants were excluded
for the following reasons: failure of the attention check (29 participants), no answers recorded
in JATOS other than a Prolific ID (10 participants), incomplete survey submission (3
participants), total completion time shorter than 90 seconds (1 participant), or total
completion time longer than 15 minutes (43 participants). This yielded 914 valid participants.
The experiment and exclusion rules were preregistered before the experiment was conducted. The
study involving human subjects was approved by Tel Aviv University\textquotesingle{}s IRB.

Despite the quality checks and initial filtering for high-quality workers, the study cannot
completely rule out the possibility that some responses obtained through Prolific might have
been generated by LLMs. If such cases occurred, they would introduce LLM-like responses into the
\textquotedbl{}human\textquotedbl{} response class and therefore work against our
method\textquotesingle{}s ability to separate the two classes, rendering the reported results
conservative. This interpretation is consistent with related work (10), which likewise notes
that if a presumed-human Prolific sample contains LLM-generated responses, the reported
detection performance estimates would be conservative.

\subsection*{LLM Data Collection and Processing}

Responses from nine LLMs at temperatures of 0.7 and 1.0 were obtained from (1): ChatGPT 3.5
Turbo, ChatGPT 4o, Claude 3.7 Sonnet, Claude 4.0 Sonnet, DeepSeek V3, Gemini 2.0, Gemini 2.5,
Grok 3, and Mistral 3. A total of 1,774 valid response instances were obtained from this
dataset.

We also included three more recent models: ChatGPT 6 Astra, Claude 5.0 Opus, and Gemini 3.6
Flash. For each of these three LLMs, 200 responses per cognitive-bias task were collected via
API. For Gemini 3.6 Flash, we collected 100 responses per prompt at each of the two temperature
values. ChatGPT 6 Astra and Claude 5.0 Opus do not support a user-defined temperature parameter,
so we collected 200 responses per prompt using their default settings.

Although all three models ultimately provided valid responses to all tasks, Gemini 3.6 Flash
declined to provide a direct response to the Egocentric task in 90\% of initial attempts and
instead returned text discussing the dilemma. Because all human responses in our sample were
valid, such non-valid responses could trivially identify a response as non-human. To make the
task more challenging and avoid trivial detection, we used repeated sampling, continuing until
200 valid direct responses to the Egocentric task were obtained. (Gemini 3.6
Flash\textquotesingle{}s responses to the other biases were all valid.)

\subsection*{Implementation of the Westwood Adversarial Survey-Taking System}

Westwood\textquotesingle{}s (11) system contains (a) a survey-platform interface that parses
Qualtrics pages and mimics human online behavior, including mouse trajectories, reading times,
and keystroke-by-keystroke typing, and (b) a core answering engine using prompt-based guidance,
persona generation, and memory of prior answers. We replicated only the answering engine because
the platform-interface layer is not a source of information used by our classifier. This design
effectively assumes flawless interaction with the survey interface by
Westwood\textquotesingle{}s system and thereby removes potential interface failures that could
otherwise make the system easier to detect.

Persona generation followed the procedure described in (11), SI section S1.1.1: independent
weighted random draws for gender, race/ethnicity, age, education, partisan affiliation, U.S.
state, and household income, with probabilities drawn from the appendix tables. Age was
generated using the two-stage procedure in (11) (age bin, then uniform age within the bin).
Because income-bracket probabilities were not provided in the appendix, the U.S. Census ACS 2022
household-income distribution over the nine brackets shown in (11) Figure 2B was used as the
closest approximation.

Two hundred personas were generated. We used the same LLM as (11), OpenAI\textquotesingle{}s
o4-mini, together with the survey-taking prompt from (11) (SI Appendix S1.1). Three JSON blocks
were injected: persona profile, accumulated prior questions and responses, and the current
question. Question order was randomized independently for each persona.
Westwood\textquotesingle{}s original prompt allowed an \textquotedbl{}I don\textquotesingle{}t
know\textquotedbl{} response, which appeared at least once for 84 personas; because none of the
human responses used that option, retaining it would have enabled trivial detection. To avoid
trivial detection, this option was therefore removed from the prompt and 200 responses were
recollected, making the evaluation more favorable to the adversarial system\textquotesingle{}s
ability to evade detection. The results may therefore be considered conservative with respect to
these implementation choices.

\subsection*{Construction of Multivariate Response Instances}

Human participants naturally provided all 11 responses within one survey session, so those
responses were stored as a single instance. Standard LLM responses, in contrast, were collected
independently for each cognitive-bias task through separate API calls; responses from a given
LLM were stacked in collection order to form complete instances. Each task was self-contained
and did not depend on preceding questions, and the classifier used only the response to each
task, not order, timing, history, or other session information. The Westwood system (11)
requires continuous sessions, so all responses from each session generated by that system were
directly assigned to a single data instance.

\subsection*{Empirical Evaluation and Feature Selection-based Ranking}

The main evaluation used a leave-one-LLM-out cross-validation-like procedure across 13
LLM/adversarial-system sets. Before the iterations, human responses were randomly shuffled and
split into 13 equal-sized parts. In each iteration, the classifier was trained on 12 LLM/system
sets and 12 human partitions, then evaluated on the held-out LLM/system responses together with
the remaining human partition. Performance was summarized by the mean area under the ROC curve
(AUC) across held-out iterations.

To evaluate shorter authentication challenges, Random Forest feature-importance scores were
calculated on each training fold. The classifier was then retrained using only the top m ranked
features selected within that training fold and applied to the corresponding held-out test fold.
The reported reduced sets use m = 5 and m = 3.

AUC was selected as the main performance measure because it does not require a prespecified
classification threshold, thereby eliminating researcher discretion in selecting an operating
threshold. Nevertheless, for robustness we also report false-positive rate (FPR) evaluated at a
specified recall level. These threshold-dependent measures are reported below as supporting data
rather than in the main text.

\subsection*{Additional Validation Procedures}

A time-based split was used to rule out the possibility that performance unfairly benefits from
training on the responses of LLMs released after the focal test LLM. Training used the nine
models whose data came from Geva et al. (2025): ChatGPT 3.5 Turbo, ChatGPT 4o, Claude 3.7
Sonnet, Claude 4.0 Sonnet, DeepSeek V3, Gemini 2.0, Gemini 2.5, Grok 3, and Mistral 3. Testing
used ChatGPT 6 Astra, Claude 5.0 Opus, and Gemini 3.6 Flash. For this analysis, human responses
were divided into 12 equal partitions; nine were used for training and three for testing.
Westwood\textquotesingle{}s system (11) was not part of the primary time split because it
combines an older o4-mini model with a newer wrapping/instruction system, so it was unclear
whether to assign it to the training or test set. For robustness, we also repeated the process
tested adding Westwood\textquotesingle{}s system to the time-based train set and obtained
similar findings (AUC = 0.994).

A second time-based split procedure also excluded same-vendor training models for each focal
held-out model. For example, when Gemini 3.6 Flash was tested, Gemini 2.0 and Gemini 2.5 were
removed from the training set. This procedure was repeated separately for ChatGPT 6 Astra,
Claude 5.0 Opus, and Gemini 3.6 Flash, and performance was averaged across the three
evaluations.

\section*{Additional Results}

\textbf{Main Analysis \textendash{} Additional Performance Measure: FPR@95\%Recall}

For the main analysis reported in the main manuscript, we detail the operating characteristics
at 95\% recall: with all 11 cognitive-bias tests, mean FPR@95\%Recall was 0.0131; with the five
highest-ranked tests, 0.0165; and with the three highest-ranked tests, 0.019.

\textbf{Validation with Time-based Split}

For the time-based split evaluation procedure (see details in \textquotedbl{}Additional
Validation Procedures\textquotedbl{} above), mean AUC was 0.991 with all 11 cognitive-bias
tests, 0.994 with the five highest-ranked tests, and 0.989 with the three highest-ranked tests.

FPR@95\%Recall for this evaluation was 0.0175 with all 11 cognitive-bias tests, 0.0175 with the
five highest-ranked tests, and 0.0132 with the three highest-ranked tests.

In the time-based, same-vendor-excluded analysis (see details in \textquotedbl{}Additional
Validation Procedures\textquotedbl{} above), mean AUC was 0.99 with all 11 cognitive-bias tests,
0.99 with the five highest-ranked tests, and 0.994 with the three highest-ranked tests.

FPR@95\%Recall for this evaluation was 0.0261 with all 11 cognitive-bias tests, 0.0319 with the
five highest-ranked tests, and 0.029 with the three highest-ranked tests.

\renewcommand{\refname}{SI References}